\documentclass{article}
\usepackage{graphicx}
\usepackage{multirow}
\usepackage{natbib}
\usepackage[linesnumbered,ruled,vlined]{algorithm2e} 
\usepackage{amsmath}

\usepackage[preprint]{neurips_2022}

\usepackage[utf8]{inputenc} 
\usepackage[T1]{fontenc}    
\usepackage{hyperref}       
\usepackage{url}            
\usepackage{booktabs}       
\usepackage{amsfonts}       
\usepackage{nicefrac}       
\usepackage{microtype}      
\usepackage{xcolor}         
\usepackage{enumitem}

\title{
CASS: Cross Architectural Self-Supervision for Medical Image Analysis}

\author{%
  Pranav Singh \\
  Department of Computer Science\\
  Tandon School of Engineering\\
  New York University
  New York, NY 11202 \\
  \texttt{ps4364@nyu.edu} \\
  \And
  Elena Sizikova \\
  Center for Data Science\\
  New York University
  New York, NY 10011 \\
  \texttt{es5223@nyu.edu} \\
  \AND
  Jacopo Cirrone \\
  Center for Data Science\\
  New York University\\
  and Colton Center for Autoimmunity\\
  NYU Grossman School of Medicine\\
  New York, NY 10011 \\
  \texttt{cirrone@courant.nyu.edu} \\
}

\begin{document}

\maketitle

\begin{abstract}
  Recent advances in deep learning and computer vision have reduced many barriers
to automated medical image analysis, allowing algorithms to process label-free
images and improve performance. However, existing techniques have extreme computational requirements and drop a lot of performance with a reduction in batch size or training epochs. This paper presents Cross Architectural - Self Supervision (CASS), a novel self-supervised learning approach that leverages Transformer and CNN simultaneously. Compared to the existing state of the art self-supervised learning approaches, we empirically show that CASS-trained CNNs and Transformers across four diverse datasets improved F1 Score and Recall value by an average of 3.8\% with 1\% labeled data, 5.9\% with 10\% labeled data, and 10.13\% with 100\% labeled data while taking 69\% less time. We also show that CASS is much more robust to changes in batch size and training epochs. Notably, one of the test datasets comprised histopathology slides of an autoimmune disease, a condition with minimal data that has been underrepresented in medical imaging. The code is open source and is available on GitHub
\end{abstract}

\section{Introduction}

In recent years, medical image analysis has seen tremendous growth due to the
availability of powerful computational modelling tools, such as neural networks, and the
advancement of techniques capable of learning from partial annotations. 
Medical imaging is a field characterized by minimal data availability. First, data labeling typically requires domain-specific knowledge. Therefore, the requirement of large-scale clinical supervision may be cost and time prohibitive. Second, due to patient privacy, disease prevalence, and other limitations, it is often difficult to release imaging datasets for secondary analysis, research, and diagnosis. Third, due to an incomplete understanding of diseases. This could be either because the disease is emerging or because no mechanism is in place to systematically collect data about the prevalence and incidence of the diseases. An example of the latter is autoimmune diseases. Statistically, autoimmune diseases affect 3\% of the US population, or 9.9 million US citizens. There are still major outstanding research questions for autoimmune diseases regarding the presence of different cell types and their role in inflammation at the tissue level. The study of autoimmune diseases is critical not only because autoimmune diseases affect a large part of society but also because these conditions have been on the rise recently \cite{galeotti2020autoimmune,lerner2015world,ehrenfeld2020covid}. Other fields like cancer and MRI image analysis have benefited from the application of artificial intelligence (AI). But for autoimmune diseases, the application of AI is particularly challenging due to minimal data availability, with the median dataset size for autoimmune diseases between 99-540 samples \cite{tsakalidou2022computer, Stafford2020ASR}. 

To overcome these limitations, we turn to self-supervised learning, a learning paradigm that allows for learning useful data representations label-free. Models extracting these representations can later be fine-tuned with a small amount of labeled data for each downstream task~\cite{sriram2021covid}. As a result, this learning approach avoids the relatively expensive and human-intensive task of data annotation and makes it an effective tool for the image analysis of emerging diseases that often have limited data availability (e.g., dermatomyositis, an autoimmune disease, or COVID-19, the cause of a recent worldwide pandemic). Existing approaches in the field of self-supervised learning rely on Convolutional Neural Networks (CNNs) or Transformers as the feature extraction backbone and learn feature representations by teaching the network to compare the extracted representations. Instead, we propose to combine a CNN and Transformer in a response-based contrastive method. In CASS, the extracted representations of each input image are compared across two branches representing each architecture (see Figure 1). By transferring features sensitive to translation equivariance and locality from CNN to Transformer, CASS learns more predictive data representations in limited data scenarios where a Transformer-only model cannot find them. We studied this qualitatively and quantitatively in Section 5. Our contributions are as follows:

\begin{itemize}
\item We introduce \textbf{C}ross \textbf{A}rchitectural  - \textbf{S}elf \textbf{S}upervision (CASS), a hybrid CNN-Transformer approach for learning improved data representations in a self-supervised setting in limited data availability problems in the medical image analysis domain 

\footnote{The code is open source and available at: 
\href{https://github.com/pranavsinghps1/CASS}{github.com/pranavsinghps1/CASS}
}.


\item We propose the use of CASS for analysis of autoimmune diseases such as dermatomyositis and demonstrate an improvement of 2.55\% 
in comparison to the existing state of the art self-supervised approaches.

\item We evaluate CASS on three challenging medical image analysis problems (autoimmune disease cell classification, brain tumor classification, and skin lesion classification) on three public datasets (Autoimmune Dataset \cite{singh2022data}, Dermofit Project Dataset \cite{Dermofit}, brain tumor MRI Dataset \cite{Cheng2017,s21062222} and ISIC 2019~\cite{Tschandl2018TheHD,Gutman2018SkinLA,Combalia2019BCN20000DL}) and find that CASS outperforms existing state of the art self-supervised techniques by an average of ~3.8\% using 1\% label fractions, 5.9 \% with 10\% label fractions and 10.13\% with 100\% label fractions (F1 Score and Recall value).

\item Existing methods also suffer a severe drop in performance when trained for a reduced number of epochs or batch size (\cite{caron2021emerging,Grill2020BootstrapYO,chen2020simple}). We show that CASS is robust to these changes in Section 5.6. 

\item New state of the art self-supervised techniques often require significant computational requirements. This is a major hurdle as these methods can take around 20 GPU days to train \cite{Azizi2021BigSM}. This makes them inaccessible in limited computational resource settings and increase triage in medical image analysis. CASS, on average, takes 69\% less time as opposed to the existing state of the art methods. We further expand on this result in Section 5.1 and further analyze this in Appendix D.
\end{itemize}
\section{Background}

\subsection{Neural Network Architectures for Image Analysis}

CNNs are a popular architecture of choice for many image analysis applications~\cite{khan2020survey}. CNNs learn more abstract visual concepts with a gradually increasing receptive field. They have two favorable inductive biases: (i) translation equivariance resulting in the ability to learn equally well with shifted object positions, and (ii) locality resulting in the ability to capture pixel-level closeness in the input data. CNNs have been used for many medical image analysis applications, such as disease diagnosis~\cite{yadav2019deep} or semantic segmentation~\cite{ronneberger2015u}. To address the requirement of additional context for a more holistic image understanding, the Vision Transformer (ViT) architecture \cite{dosovitskiy2020image} has been adapted to images from language-related tasks and recently gained popularity~\cite{liu2021Swin,liu2021swinv2,pmlr-v139-touvron21a}. In a ViT, the input image is split into patches, that are treated as tokens in a self-attention mechanism. In comparison to CNNs, ViTs can capture additional image context, but lack ingrained inductive biases of translation and location. As a result, ViTs typically outperform CNNs on larger datasets~\cite{d2021convit}. 

ConViT~\cite{d2021convit} combines CNNs and ViTs using gated positional self-attention (GPSA) to create a soft-convolution similar to inductive bias and improve upon the
capabilities of Transformers alone. More recently, the training regimes and inferences from ViTs have been used to design a new family of convolutional architectures - ConvNext \cite{liu2022convnet}, outperforming benchmarks set by ViTs in classification tasks.

\subsection{Self-Supervised Learning for Medical Imaging}
Self-supervised learning allows for the learning of useful data representations without data labels~\cite{NEURIPS2020_f3ada80d}, and is particularly attractive for medical image analysis applications where data labels are difficult to find~\cite{Azizi2021BigSM}. Recent developments have made it possible for self-supervised methods to match and improve upon existing supervised learning methods~\cite{hendrycks2019using}.

However, existing self-supervised techniques typically require large batch sizes and datasets. When these conditions are not met, a marked reduction in performance is demonstrated~\cite{caron2021emerging,chen2020simple,Caron2020UnsupervisedLO,Grill2020BootstrapYO}. Self-supervised learning approaches have been shown to be useful in big data medical applications~\cite{ghesu2022self,azizi2021big}, such as analysis of dermatology and radiology imaging. In more limited data scenarios (3,662 images - 25,333 images), \citet{Matsoukas2021IsIT} reported that ViTs outperform their CNN counterparts when self-supervised pre-training is followed by supervised fine-tuning. Transfer learning favors ViTs when applying standard
training protocols and settings. Their study included running the DINO \cite{caron2021emerging} self-supervised method over 300 epochs with a batch size of 256. However, questions remain about the accuracy and
the efficiency of using existing self-supervised techniques when using them on datasets whose entire size is smaller than their peak performance batch size. Also, viewing this from the general practitioner's perspective with limited computational power raises the
question of how we can make practical self-supervised approaches more accessible? Adoption and faster development of self-supervised paradigms will only be possible when they become easy to plug-and-play with limited computational power.

In this work, we explore these questions by designing CASS, a novel approach developed with the core values of efficiency and effectiveness. In simple terms, we are combining CNN and Transformer in a response-based contrastive method by reducing similarity to combine the abilities of CNNs and Transformers. This approach was originally designed for a 198 image dataset for muscle biopsies of inflammatory lesions from patients who have dermatomyositis - an autoimmune disease. The benefits of this approach are illustrated by challenges in the diagnosis of autoimmune diseases due to their rarity, limited data availability, and heterogeneous features. As a consequence, misdiagnoses are common, and the resulting diagnostic delay plays a major factor in their high mortality rate. Autoimmune diseases also share commonalities with COVID-19 in terms of clinical manifestations, immune responses and pathogenic mechanisms. Moreover, some patients have developed autoimmune diseases after COVID-19 infection \cite{Liu2020COVID19AA}. Despite this increasing prevalence, the representation of autoimmune diseases in medical imaging and deep learning is limited. Furthermore, developing effective and efficient techniques such as CASS will aid in their widespread adoption, further expanding the work in multiple domains and resulting in a  multi-fold improvement in quality of life.

Recent self-supervised methods define the inputs as two augmentations of one image and maximize the similarity between the two representations, by passing them through a pair of feature extractors. These feature extractors are similar in structure and only differ in their weights/parameters. Methods like Momentum Contrast (MoCo) \cite {He2020MomentumCF} and SiMCLR \cite{Chen2020ASF} maintain negative samples in a memory queue. The core idea in such scenarios is to bring the positive pairs together while repulsing the
negative sample pairs. Recently, Bootstrap Your Own Latent (BYOL) \cite{Grill2020BootstrapYO} and DINO \cite{caron2021emerging} have improved upon this approach by eliminating the memory banks. The premise of using negative pairs is to avoid collapse. Several strategies have been developed with BYOL using a momentum encoder, Simple Siamese (SimSiam) \cite{Chen2021ExploringSS} a stop gradient, and DINO applying the counterbalancing effects of sharpening and centering to avoid collapse. DINO is the first self-supervised training approach extended for Transformers. As described on the right side of Figure 1, DINO augments an image to produce two versions of the image; these are then passed through the student and teacher networks, which are essentially the same encoder with different parameters. Their similarity is then measured with a cross-entropy loss. A stop-gradient (sg) operator is applied to the teacher network to propagate gradients only through the student network.

\section{Methodology}

We start by motivating our method before explaining in detail (in Section 3.1). Self-supervised methods until now have been using different augmentations of the same image to create positive pairs. These were then passed through same architectures but with different set of parameters \cite{Grill2020BootstrapYO}. \\
In \cite{caron2021emerging} they introduced image cropping of different sizes to add local and global information. They also used different operators and techniques to avoid collapse as described in Section 2.2. \\

\cite{Raghu2021DoVT} in their study suggested that for the same input, Transformers and CNNs extract different representations. They conducted their study by analyzing the CKA (Centered Kernel Alignment) for CNNs and Transformer using ResNet \cite{He2016DeepRL} and ViT (Vision Transformer) \cite{dosovitskiy2020image} family of encoders respectively. They found that Transformers have a more uniform representation across all layers as compared to CNNs. They also have self-attention, enabling global information aggregation from shallow layers and skip connections that connect lower layers to higher layers, promising information transfer. Hence, lower and higher layers in Transformers show much more similarity than in CNNs. The receptive field of lower layers for Transformers is more extensive than in CNNs. While this receptive field gradually grows for CNNs, it becomes global for Transformers around the midway point. Transformers don't attend locally in their earlier layers, while CNNs do. Using local information earlier is important for strong performance. CNNs have a more centered receptive field as opposed to a more globally spread receptive field of Transformers. Hence, representations drawn from the same input, will be different for Transformers and CNNs. Until now self-supervised techniques have used only one kind of architecture either a CNN or Transformer. But differences in the representations learned with CNN and Transformers, inspired us to create positive pairs by different architectures or feature extractors rather than using different set of augmentations. This by design avoids collapse as the two architectures will never give the same representation as output. By contrasting their extracted features at the end we hope to help the Transformer learn representations from CNN and vice versa.
This should help both the architectures to learn better representations. We verify this, by studying attention maps and feature maps from supervised and CASS trained CNN and Transformers in Appendix D. We observed that CASS trained CNN and Transformer were able to retain a lot more detail about the input image which pure CNN and Transformers lacked. Furthermore, cross-architecture knowledge distilled models have shown encouraging improvements over supervised-knowledge distillation \cite{Gong2022CMKDCC}.

\subsection{Description of CASS}

CASS' goal is to extract and learn representations in a self-supervised way. To achieve this, an image is passed through a common set of augmentations. The augmented image is then simultaneously passed though a CNN and Transformer to create positive pairs. The output logits from the CNN and Transformer are then used to find cosine similarity loss (equation \ref{loss_eq}). Since, the two architectures give different output representations as mentioned in \cite{Raghu2021DoVT}, the model doesn't collapse. We also report results for CASS using different set of CNNs and Transformers in Appendix B and C, and not a single case of model collapse was registered.

\begin{equation}
\label{loss_eq}
\operatorname{loss} =2-2 \times
 \operatorname{F(R)} \times \operatorname{F(T)}
 \end{equation}
\begin{align*}
\text{where, }
\operatorname{F(x)}=\sum_{i=1}^{N} \left(\frac{x}{\left(\operatorname{max}\left(\|x\|_{2}\right), \epsilon\right)}\right)
\end{align*}

We use same parameters for optimizer and learning schedule for both the architectures. We also use stochastic weigh averaging (SWA) \cite{Izmailov2018AveragingWL} with Adam optimizer and a learning rate of 1e-3. For learning rate we use cosine schedule with a maximum of 16 iterations and a minimum value of 1e-6. ResNets are typically trained with Stochastic Gradient Descent (SGD) and our use of Adam optimizer is quite unconventional. Furthermore, unlike existing self-supervised techniques there is no parameter sharing between the two architectures. 

In Figure 1, we show CASS on top and DINO \cite{caron2021emerging} at the bottom. Comparing the two, CASS does not use any extra mathematical treatment used in DINO to avoid collapse such as centring and applying softmax function on the output of its student and teacher networks. After training CASS and DINO for one cycle, DINO yields only one kind of trained architecture, while CASS provides two trained architectures (1 - CNN and 1 - Transformer). CASS trained architectures provide better performance than DINO trained architectures in most cases as further elaborated in Section 5. 

\begin{figure}[!htb]
    \centering
    \label{fig:CASS_fig}
    \centering  
    \includegraphics[width=0.7\linewidth]{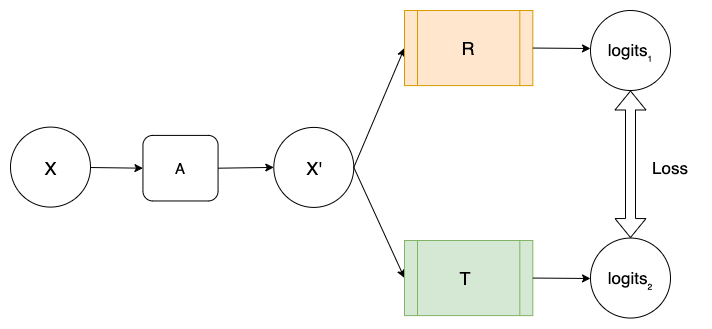}
    \hspace{1cm}
    \includegraphics[width=0.7\linewidth]{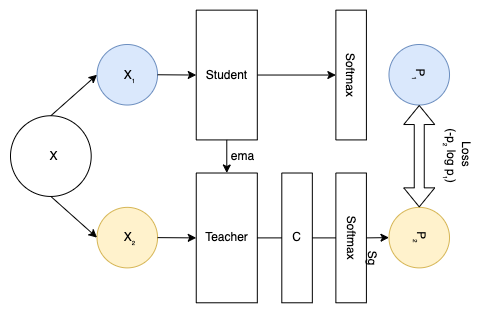}
    \caption{(Top) In our proposed self-supervised architecture - CASS, R represents ResNet-50, a CNN and T in the other box represents the Transformer used (ViT); X is the input image, which becomes X' after applying augmentations. Note that CASS applies only one set of augmentations to create X'. X' is then passed through both the arms to compute loss as mentioned in Equation 1. This is different from DINO, which passes different augmentation of the same image through networks with the same architecture but different parameters. The output of the teacher network is centred on a mean computed over a batch. Another key difference is that in CASS, loss is computed over logits meanwhile in DINO it is computed over softmax output.}
\end{figure}
\section{Experimental Details}

\subsection{Self-supervised learning}
We studied and compared results between DINO and CASS trained self-supervised CNNs and Transformers. For the same, we trained from ImageNet initialization for 100 epochs with a batch size of 16. We ran these experiments on an internal cluster with single GPU unit (NVIDIA RTX8000) with 48 GB video RAM, 2 CPU cores and 64 GB system RAM. 

For DINO, we used the hyper parameters and augmentations mentioned in the original implementation. \\
For CASS, we describe the experimentation details in Appendix E.

\subsection{End-to-end fine-tuning} In order to evaluate the utility of the learned representations, we use the self-supervised pre-trained weights for downstream classification task. While performing the downstream fine tuning we perform entire model (E2E
fine-tuning). The test set metrics were used as proxies for representation quality. We trained the entire model for a maximum of 50 epochs with an early stopping patience of 5 epochs. For supervised fine tuning we used Adam optimizer with a cosine annealing learning rate starting at 3e-04. Since almost all medical datasets have some class imbalance we applied class distribution normalized Focal Loss \cite{Lin2017FocalLF} to navigate class imbalance. 
We fine tune the models with different label fractions during training i.e 1\%, 10\% and 100\% label fractions. For example, if a model is trained with 10\% label fraction then that model will have access only to 10\% of the training dataset labels, which was used during self-supervised training. 

\subsection{Datasets}

We split the datasets into three splits - training, validation and testing following the 70/10/20 split strategy unless specified otherwise. We futher expand upon our thought process for choosing datasets in Appendix E.

\begin{itemize}

    \item \textbf{Autoimmune diseases biopsy slides} (\cite{VANBUREN2022113233,singh2022data}) consists of slides cut from muscle biopsies of dermatomyositis patients stained with different proteins and imaged to generate a dataset of 198 TIFF image set from 7 patients. The presence or absence of these cells helps to diagnose dermatomyositis. Multiple cell classes can be present per image; therefore this is a multi-label classification problem. 
    Our task here was to classify cells based on their protein staining into TFH-1, TFH-217, TFH-Like, B cells, and others. We used F1 score as our metric for evaluation, as employed in previous works by \cite{singh2022data, VANBUREN2022113233}. These RGB images have a consistent size of 352 by 469.

    \item \textbf{Dermofit dataset} \cite{Dermofit}  contains normal RGB images captured through SLR camera indoors with ring lightning. There are 1300 image samples, classified into 10 classes: Actinic Keratosis (AK), Basal Cell Carcinoma (BCC), Melanocytic Nevus / Mole (ML), Squamous Cell Carcinoma (SCC), Seborrhoeic Keratosis (SK), Intraepithelial carcinoma (IEC), Pyogenic Granuloma (PYO), Haemangioma (VASC), Dermatofibroma (DF) and  Melanoma (MEL). This dataset comprises of images of different sizes and no two images are of same size. They range from 205×205 to 1020×1020 in size. We pretext task is multi-class classification and we use F1 score as our evaluation metric on this dataset.

    \item \textbf{Brain tumor MRI dataset} \cite{Cheng2017,Amin2022ANM}  7022 images of human brain MRI that are classified into four classes: glioma, meningioma, no tumor, and pituitary. We used the dataset from \url{https://www.kaggle.com/datasets/masoudnickparvar/brain-tumor-mri-dataset} that combines Br35H: Brain tumor Detection 2020 dataset used in "Retrieval of Brain tumors by Adaptive Spatial Pooling and Fisher Vector Representation" and Brain tumor classification curated by Navoneel Chakrabarty and Swati Kanchan. Out of these, the dataset curator created the training and testing splits. We followed their splits, 5,712 images for training and 1,310 for testing. Since this was a combination of multiple datasets, size of images vary throughout the dataset from 512×512 to 219×234. The pretext of the task is multi-class classification, and we used the F1 score as the metric.

    \item \textbf{ISIC 2019} (\cite{Tschandl2018TheHD,Gutman2018SkinLA,Combalia2019BCN20000DL}) consists of 25,331 images across eight different categories - melanoma (MEL), melanocytic nevus (NV), Basal cell carcinoma (BCC), actinic keratosis(AK), benign keratosis(BKL) , dermatofibroma(DF), vascular lesion (VASC) and Squamous cell carcinoma(SCC). This dataset contains images of size 600 × 450 and 1024 × 1024. The distribution of these labels is unbalanced across different classes. For evaluation, we followed the metric followed in the official competition i.e balanced multi-class accuracy value, which is semantically equal to recall. 
    \end{itemize}

\section{Results and Discussion}

\subsection{Compute and Time analysis Analysis}

We ran all the experiments on a single NVIDIA
RTX8000 GPU with 48GB video memory. In Table \ref{computetime}, we compare the cumulative training times for self-supervised training of a CNN and Transformer with DINO and CASS. We observed that CASS took an average of 69\% less time compared to DINO. Another point to note is that, CASS trained two architectures at the same time or in a single pass. While to train a CNN and Transformer with DINO it would take two separate passes.

\begin{table}[!htb]
\centering
\begin{tabular}{lll}
\hline
Dataset    & DINO              & CASS                    \\
\hline
Autoimmune & 1 Hour 13 Mins    & \textbf{21 Mins}          \\
Dermofit & 3 Hours 9 mins & \textbf{1 Hour 11 Mins} \\
Brain MRI  & 26 Hours 21 Mins  & \textbf{7 Hours 11 Mins}  \\
ISIC-2019  & 109 Hours 21 Mins & \textbf{29 Hours 58 Mins} \\
\hline
\end{tabular}
\caption{Self-supervised training time comparison for 100 epochs on a single RTX8000 GPU.}
\label{computetime}
\end{table}

\subsection{Autoimmune Diseases Biopsy Slides Dataset}
We did not perform 1\% training for the autoimmune diseases biopsy slides of 198 images because using 1\% images would be too small number to learn anything meaningful and the results would be highly randomized.

Following the self-supervised training and fine tuning procedure as described in Section 4.1 and 4.2, we observed that using CASS with the ViT B/16 backbone and ResNet50 improved upon existing state of the art results from DINO trained CNN and Transformers. 


\begin{table}[!htb]
\centering
\begin{tabular}{llll}
\hline
\multicolumn{1}{c}{\multirow{2}{*}{Techniques}} & \multicolumn{1}{c}{\multirow{2}{*}{Backbone}} & \multicolumn{2}{l}{Testing F1 score} \\
\multicolumn{1}{c}{}                            & \multicolumn{1}{c}{}                           & 10\%       & 100\%         \\
\hline

DINO                                            & Resnet-50                                      & \textbf{0.8237±0.001}            &0.84252±0.008           \\
CASS                                           & Resnet-50                                   & 0.8158±0.0055           & \textbf{0.8650±0.0001}           \\
Supervised                                      & Resnet-50                                 & 0.819±0.0216          &  0.83895±0.007            \\
\hline

DINO                                            & ViT B/16                             & 0.8445±0.0008           &0.8639± 0.002             \\
CASS                                           & ViT B/16                                          & \textbf{0.8717±0.005}           & \textbf{0.8894±0.005}            \\
Supervised                                      & ViT B/16                                          &0.8356±0.007           &0.8420±0.009          \\
\hline
\end{tabular}
\caption{Results for autoimmune biopsy slides dataset. In this table we compare the F1 score on test set. We observed that CASS outperformed the existing state-of-art self-supervised method using 100\% labels for CNN as well as for Transformers. Although DINO outperforms CASS for CNN with 10\% labeled fraction. Overall CASS outperforms DINO by ~2.2\% for 100\% labeled training for CNN and Transformer. For Transformers in 10\% labeled training CASS' performance was ~2.7\% better than DINO.}
\end{table}

\subsection{Dermofit Dataset}
We did not perform 1\% training for this dataset as the training set was too small to draw meaningful results with just 10 samples. We observe that CASS outperforms both supervised and existing state of the art self-supervised methods for all label fractions. We present the F1 score for different label fractions in Table \ref{dermofitperformance}. 
\begin{table}[!htb]
\centering
\begin{tabular}{lll}

\hline
\multicolumn{1}{c}{\multirow{2}{*}{Techniques}} & 
\multicolumn{2}{l}{Testing F1 score} \\
\multicolumn{1}{c}{}                                         & 10\%       & 100\%         \\
\hline

DINO                                             (Resnet-50)                                     &0.3749±0.0011          &0.6775±0.0005           \\
CASS                                            (Resnet-50)                                     & \textbf{0.4367±0.0002}      & \textbf{0.7132±0.0003}           \\
Supervised                                       (Resnet-50)                                   & 0.33±0.0001           &  0.6341±0.0077           \\
\hline

DINO                        (ViT B/16)                             &0.332± 0.0002             &0.4810±0.0012            \\
CASS    (ViT B/16)                                  & \textbf{0.3896±0.0013}           & \textbf{0.6667±0.0002}            \\
Supervised         (ViT B/16)                               &0.299±0.002          &0.456±0.0077          \\
\hline
\end{tabular}
\caption{Results for the dermofit dataset. Parenthesis next to the techniques represent the architecture used, for example DINO(ViT B/16) represents ViT B/16 trianed with DINO. In this table we compare the F1 score on test set. We observed that CASS outperformed the existing state-of-art self-supervised method using for all label fractions and for both the architectures.}
\label{dermofitperformance}
\end{table}

\subsection{Brain tumor MRI dataset}

We observed that supervised CNNs performed better than Transformers on this dataset. Similarly, for DINO, CNNs performed better than Transformers by a margin. We observed that this trend is followed by CASS as well. The difference between CNN and Transformer performance is smaller for CASS as compared to the difference in performance for CNN and Transformer with supervised and DINO training. We report these results in Table \ref{brainMRIperformance}.
\begin{table*}[t]
\centering
\begin{tabular}{lllll}
\hline
\multicolumn{1}{c}{\multirow{2}{*}{Techniques}} & \multicolumn{1}{c}{\multirow{2}{*}{Backbone}} & \multicolumn{3}{l}{Testing F1 score} \\
\multicolumn{1}{c}{}                            & \multicolumn{1}{c}{}                           & 1\%       & 10\%       & 100\%       \\
\hline
DINO                                            & Resnet-50                                      &\textbf{0.63405±0.09}          &\textbf{0.92325±0.02819}            & 0.9900±0.0058          \\
CASS                                           & Resnet-50                                      & 0.40816±0.13          & 0.8925±0.0254          &\textbf{0.9909±
0.0032}
             \\
Supervised                                      & Resnet-50                                      &0.52±0.018          &0.9022±0.011            & 0.9899± 0.003            \\
\hline
DINO                                            & ViT B/16                                          &0.3211±0.071      &0.7529±0.044           &0.8841±
0.0052
           \\
CASS                                           & ViT B/16                                           & \textbf{0.3345±0.11}          & \textbf{0.7833±0.0259}           &\textbf{0.9279± 
0.0213}
             \\
Supervised                                      & ViT B/16                                           & 0.3017 ± 0.077         & 0.747±0.0245           & 0.8719± 0.017           \\
\hline
\end{tabular}
\caption{While DINO outperformed CASS for 1\% and 10\% labeled training for CNN, CASS maintained its superiority for 100\% labeled training, albeit by just 0.09\%. Similarly, CASS outperformed DINO for all data regimes for Transformers, incrementally 1.34\% in for 1\%, 3.04\% for 10\%, and 4.38\% for 100\% labeled training. We observe that this margin is more significant than for biopsy images. Such results could be ascribed to the increase in dataset size and increasing learnable information.}
\label{brainMRIperformance}
\end{table*}

\subsection {ISIC 2019 Dataset}

The ISIC-2019 dataset is an incredibly challenging dataset, not only because of the class imbalance issue but because it is made of partially processed and inconsistent images with hard-to-classify classes. From Table \ref{ISICperformance} it is clear that CASS outperforms DINO for all label fractions for both CNN and Transformer by a margin.
\begin{table*}[t]
\centering
\begin{tabular}{lllll}
\hline
\multicolumn{1}{c}{\multirow{2}{*}{Techniques}} & \multicolumn{1}{c}{\multirow{2}{*}{Backbone}} & \multicolumn{3}{l}{Testing Balanced multi-class accuracy} \\
\multicolumn{1}{c}{}                            & \multicolumn{1}{c}{}                           & 1\%       & 10\%       & 100\%       \\
\hline
DINO                                            & Resnet-50                                      &0.328±0.0016         &0.3797±0.0027            &0.493±3.9e-05            \\
CASS                                           & Resnet-50                                      &\textbf{0.3617±0.0047}            &\textbf{0.41±0.0019}            &  \textbf{0.543±2.85e-05}           \\
Supervised                                      & Resnet-50                                      &0.2640±0.031           &0.3070±0.0121            &0.35±0.006           \\ 
\hline
DINO                                            & ViT B/16                                           & 0.3676± 0.012           &0.3998±0.056          &0.5408±0.001            \\
CASS                                           & ViT B/16                                           &\textbf{0.3973± 0.0465}           &\textbf{0.4395±0.0179}            &  \textbf{0.5819±0.0015}          \\
Supervised                                      & ViT B/16                                           &0.3074±0.0005           & 0.3586±0.0314           &  0.42±0.007  \\
\hline
\end{tabular}

\caption{Results for the ISIC-2019  dataset. Comparable to the official metrics used in the challenge \url{https://challenge.isic-archive.com/landing/2019/}. We use balanced multi-class accuracy as our metric, which is semantically equal to recall value. We observed that CASS consistently outperforms DINO by approximately 4\% for all label fractions with CNN and Transformer.}
\label{ISICperformance}
\end{table*}

\subsection{Ablation Studies}

\subsubsection{Change in batch size}
To gauge the change in performance with the change in batch size, we ran experiments with CASS with a batch size of 8, 16, and 32. We present these results in Appendix B.1 and C.1. Based on them we concluded that CASS is robust to change in batch size. Interestingly, instead of dropping performance like existing methods, CASS-trained Transformers with smaller batch sizes performed better. 

\subsubsection{Training Epochs} We report the performance of CASS trained for 50, 100, 200 and 300 epochs in Appendix B.2 and C.2. On reducing the epochs to 50, a performance drop of 2\% was observed for CNN, while the performance of Transformer remained almost constant. Similarly, there is a ~2\% gain when we double the number of self-supervised training epochs. However, after that, the gain is minimal on changing epochs from 200 to 300. 

\subsubsection{Effect of augmentation change} CASS does not use hard augmentations like DINO or BYOL.
We study the effect of adding BYOL/DINO-like augmentations in Appendix B.3. Although, Gaussian blur helps in converging the CNN and Transformer for CASS. We find that adding BYOL-like hard augmentations costs performance. CASS has global-local cropping inbuilt due to difference in the receptive field of CNN and Transformer, unlike DINO, where it was added with augmentation. 

\subsubsection{Change in architecture} We provide intuition for changing the architecture of CNNs and Transformers in Appendix B.5 and C.3. As a baseline we started with ResNet-50 and ViT Base/16 Transformer for our experiments. But we also expand these results to other CNN and Transformer families. Furthermore, we also report results of using two CNNs or two Transformers on the brain MRI classification dataset in Appendix C.3.

\subsubsection{Optimization} For CASS, we used Adam optimizer for both CNN and Transformer. Traditionally, CNNs and more specifically ResNets have been used with a SGD optimizer, but in our case we use Adam optimiser. This choice is fairly unconventional and we further expand upon this in Appendix B.


\subsubsection{Studying the attention and feature maps}Our motivation to combine CNN and Transformer was to help the two architectures learn meaningful representations from each other. As mentioned already in Section 3, the two architectures focus on different parts of the image and hence create positive pairs without differential augmentation. In this section we study the feature maps and attention maps of CNN and Transformer respectively to see this gain qualitatively. Quantitative gains have been summarized in Section 5. We present the attention and feature maps for a given input image \ref{fig:main_sample_image} in this section and expand the study of feature and attention maps in Appendix D.

\begin{figure}
    \centering
    \includegraphics[width=0.6\linewidth]{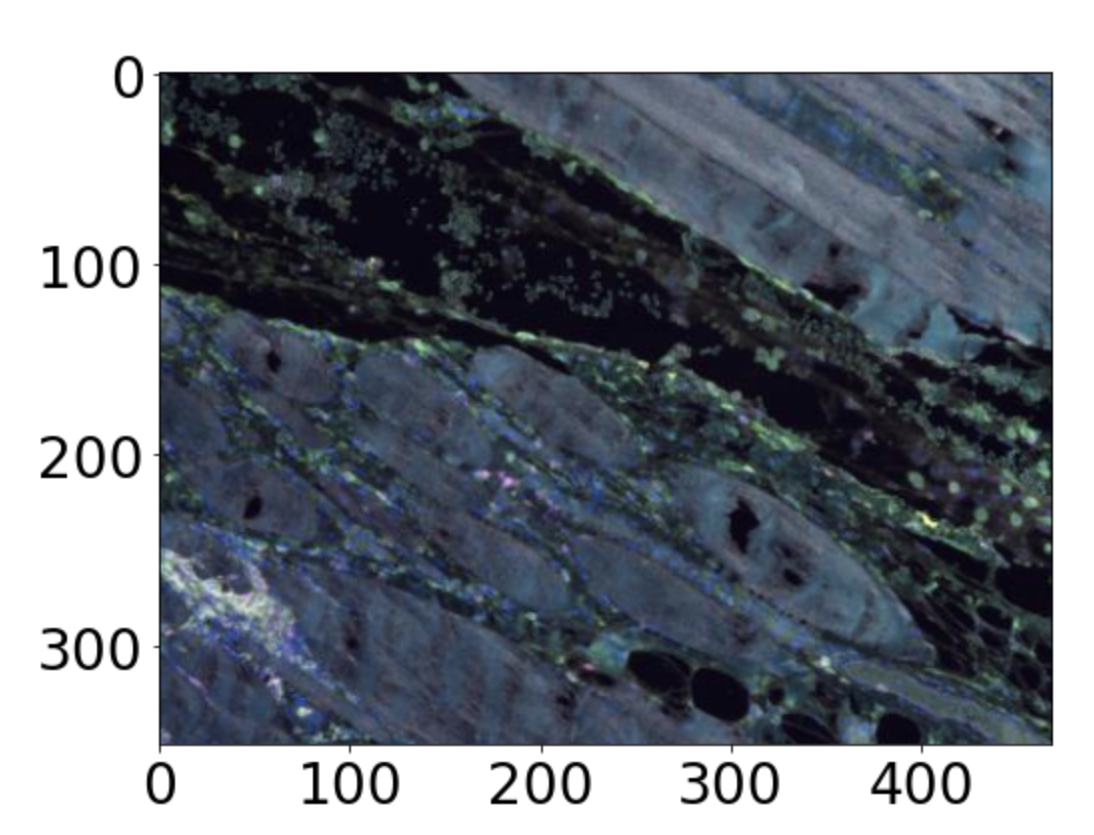}
    \caption{Sample image used from the test set of the autoimmune dataset.}
    \label{fig:main_sample_image}
\end{figure}
\begin{figure}
    \centering
    \includegraphics[width=0.45\linewidth]{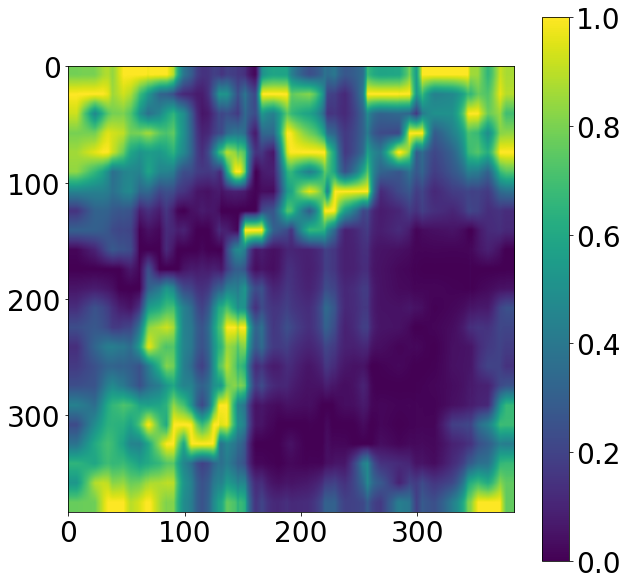}
    \includegraphics[width=0.45\linewidth]{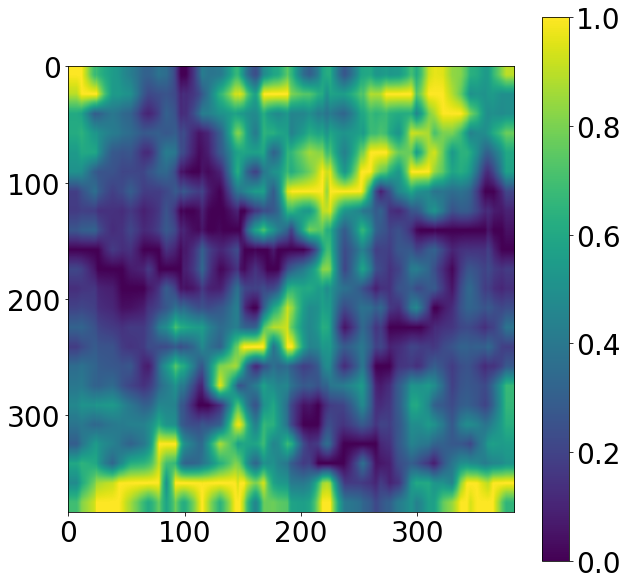}
    \caption{Overall attention maps from supervised Transformer (on the left) and CASS trained Transformer (on the right). We pass the same image as input through both of them; the image used is shown in Figure \ref{fig:main_sample_image}. We observed that the CASS-trained Transformer's attention is more spread as compared to supervised Transformer. This can be easily inferred from the right-hand side bottom portion of both the attention maps.}
    
    \label{fig:attnmapssingle}
\end{figure}
\begin{figure}[!ht]
    \centering
    \includegraphics[width=1\linewidth]{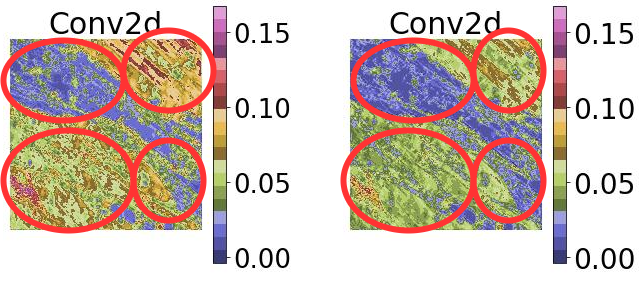}
    \caption{This figure shows the feature map extracted after the first layer of ResNet-50 for CASS (on the left) and supervised CNN (on the right). From the four circles, it is clear that CASS-trained CNN can retain much more intricate detail about the input image (Figure \ref{fig:main_sample_image}) that the supervised CNN misses.}
    
    \label{fig:fmapssingle}
\end{figure}

\begin{itemize}

\item \textbf{Feature maps.} We studied the feature maps after Conv2d layer of ResNet-50. In this section, we focus on the features extracted after the first layer of ResNet-50 trained with CASS and supervised technique in Appendix D.2. We observed that CASS trained Resnet-50 was able to retain a lot more detail/information about the input as compared to the supervised ResNet-50. We present the extracted features in Figure \ref{fig:fmapssingle}. Furthermore, we expand this study to include feature maps from the first five layers of CASS and supervised ResNet-50 in Appendix D.

\item \textbf{Class attention maps.} Similar to feature maps of CNNs, for Transformers we study their class attention maps. We extracted the attention maps after the first attention block. For the sample image in \ref{fig:main_sample_image}, we present the attnetion maps resized to the image size in \ref{fig:attnmapssingle}. We observed that CASS trained Transformer is able to pay more attention to the intricate details in the input image as compared to the supervised Transformer. Furthermore, It is able to pay attention to areas that are missed by a supervised Transformer. We also expand this study to study attention maps averaged over 30 samples on different datasets in Appendix D. 

\end{itemize}

\section{Limitations}

Although CASS' performance for larger and non-biological data can be hypothesized based on inferences, a complete study on large-sized natural datasets hasn't been conducted. In this study, we focused extensively on studying the effects and performance of our proposed method for small dataset sizes and limited computational resources. Furthermore, all the datasets used in our experimentation are restricted to academic and research use only. Although CASS performs better than existing self-supervised and supervised techniques, it is impossible to determine at inference time (without ground-truth labels) whether to pick the CNN or the Transformers arm of CASS.

\section{Potential negative societal impact}

The autoimmune dataset is limited to a geographic institution. Hence the study is specific to a disease variant. Inferences drawn may or may not hold true for other variants.
Also, the results produced are dependent on a set of markers. Medical practitioners often require multiple tests before finalising diagnosis; medical history and existing health conditions also play an essential role. We haven't incorporated the aforementioned meta-data in CASS. Finally, application on a wider scale - real life scenarios should only be trusted after taking clearance form the concerned health and safety governing bodies.

\section{Conclusion}

Based on our experimentation on four diverse medical imaging datasets, we qualitatively and empirically conclude that CASS gained an average of 3.8\% with
1\% labeled data, 5.9\% with 10\% labeled data, and 10.13\% with 100\% labeled data and trained in 69\% less time than the existing state of the art self-supervised method. Furthermore, we saw that CASS is robust to batch size changes and training epochs reduction.
To conclude, for medical image analysis, CASS is computationally efficient, performs better, and overcomes some of the shortcomings of existing self-supervised techniques. This ease of accessibility and better performance will catalyze medical imaging research to help us improve healthcare solutions and develop new solutions for underrepresented and emerging diseases.

\bibliography{ref}


\appendix

\section{Appendix}

\newcommand\mycommfont[1]{\footnotesize\ttfamily\textcolor{blue}{#1}}
\SetCommentSty{mycommfont}

\SetKwInput{KwInput}{Input}                
\SetKwInput{KwOutput}{Output}  

\section{Self-supervised Algorithm}

The core self-supervised algorithm, used to train CASS with a CNN (R) and a Transformer (T), is described in Algorithm \ref{alg:algorithm}. Here, num\_epochs represents the number of self-supervised epochs to run. CNN and Transformer represents the respective architecture we use, for example CNN could be a ResNet50 and Transformer can be ViT Base/16.
The Loss used in line 5, is described in Equation \ref{loss_eq}. Finally, after training for downstream fine tuning we save the CNN and Transformer at the defined 'path' mentioned in lines 12 and 13.
\begin{algorithm}[!ht]
\caption{Herein we describe CASS self-supervised training algorithm}  
\label{alg:algorithm}
\DontPrintSemicolon
  
  \KwInput{Unlabeled same augmented images from the training set $x'$}
  \KwOutput{Logits from each network.}
  \KwData{Images from a given dataset}
  \For{x in train loader:}
  {$R = cnn( x')$
  \tcp{taking logits output from CNN}
  $T= vit(x')$
  \tcp{taking logits output from ViT}
  $
  \operatorname{loss} = 2-2^{*}\left(\sum_{i=1}^{N}\left(\frac{R}{\left(\operatorname{man}\left(\|R\|_{2}\right), \epsilon\right)}\right) \times \sum_{i=1}^{N}\left(\frac{T}{\left.\operatorname{man}\left(\|T\|_{2}, \epsilon\right)\right)}\right)\right.
  $
  \tcp{taking cosine similarity between the logits outputs from CNN and ViT}
  Calculate the mean value of all elements of the loss tensor.
  
  Compute gradients.}
\end{algorithm}

\section{Ablation Study}

\subsection{Batch size}

As mentioned, with CASS, we aim to overcome the shortcomings of existing self-supervised methods where they drop much performance with a reduction in batch size. To study this effect, we ran CASS with three different batch sizes on the autoimmune dataset - 8, 16, and 32 and reported the results in \ref{batchsize_atm}. We observed that the performance of CASS-trained CNN improved with a reduction in batch size while that of the Transformer remained almost constant. We followed the standard set of protocols mentioned in Appendix E for training. As standard, we used 16 as our batch size. We also conducted this experiment on the brain MRI classification dataset and reported the results in Appendix C.1.

\begin{table}[!htb]
\centering
\begin{tabular}{cll}
\hline
\multicolumn{1}{l}{Batch Size} & CNN F1 Score  & Transformer F1 Score \\
\hline
8                              & 0.88285±0011  & 0.8844±0.0009        \\
16                             & 0.8650±0.0001 & 0.8894±0.005         \\
32                             & 0.8648±0.0005 & 0.889±0.0064       \\
\hline
\end{tabular}
\caption{F1 metric comparison between the two arms of CASS trained over 100 epochs, following the protocols and procedure listed in Appendix E. We only change the batch size during self-supervised training. Based on these we observed that while CASS trained Transformer gains 1\% with reduction in batch size from 16 to 8, CASS trained CNN gains almost ~2\% for the same change. Although there is a diminishing gain in performance as we increase the batch size.}
\label{batchsize_atm}
\end{table}

\subsection{Change in training epochs}

As standard, we trained CASS for 100 epochs in all cases. However, existing self-supervised techniques are plagued with a loss in performance with a decrease in the number of training epochs. To test this effect for CASS, we ran it for 50, 100, 200, and 300 epochs on the autoimmune dataset and reported the results in Table \ref{epochs_atm}. We observed a slight gain in performance when we increased the epochs from 100 to 200 but minimal gain beyond that. We also ran the same experiment on the brain MRI dataset and reported the results in Appendix C.2. 

\begin{table}[!htb]
\centering
\begin{tabular}{lll}
\hline
\multicolumn{1}{c}{Epochs} & CNN F1 Score  & Transformer F1 Score \\
\hline
50                         & 0.8521±0.0007 & 0.8765± 0.0021       \\
100                        & 0.8650±0.0001 & 0.8894±0.005         \\
200                        & 0.8766±0.001  & 0.9053±0.008         \\
300                        & 0.8777±0.004  & 0.9091±8.2e-5       \\
\hline
\end{tabular}
\caption{ Performance comparison over varied number of epochs, from 50 to 300 epochs, the downstream training procedure and the CNN-Transformer combination is kept constant across all the four experiments, only the number of self-supervised epochs have been changed.}
\label{epochs_atm}
\end{table}

\subsection{Augmentations}

Contrastive learning techniques are known to be highly dependent on augmentations. Recently, most self-supervised techniques have adopted BYOL \cite{Grill2020BootstrapYO}-like a set of augmentations. DINO \cite{caron2021emerging} uses the same set of augmentations as BYOL, along with adding local-global cropping. We use a reduced set of BYOL augmentations for CASS, along with a few changes. For instance, we do not use solarize and Gaussian blur. Instead, we use affine transformations and random perspectives. In this section, we study the effect of adding BYOL-like augmentations to CASS. We report these results in Table \ref{augmentaions}. We observed CASS trained CNN is robust to changes in augmentations. On the other hand, the Transformer drops performance with change in augmentations. A possible solution to regain this loss in performance for Transformer with change in augmentation is by using Gaussian blur, which has a converging effect on the results of CNN and Transformer. 

\begin{table*}[!htb]
\label{table:augchnages}
\centering
\begin{tabular}{cll}
\hline
Augmentation Set                                    & CNN F1 Score    & Transformer F1 Score \\
\hline
CASS only                                           & 0.8650±0.0001   & 0.8894±0.005         \\
CASS + Solarize                                     & 0.8551±0.0004   & 0.81455±0.002        \\
CASS + Gaussian blur                                & 0.864±4.2e-05   & 0.8604±0.0029        \\

\multicolumn{1}{l}{CASS + Gaussian blur + Solarize} & 0.8573±2.59e-05 & 0.8513±0.0066       \\
\hline
\end{tabular}
\caption{We report the F1 metric of CASS trained with
a different set of augmentations for 100 epochs. While CASS-trained CNN fluctuates within a percent of its peak performance, CASS-trained Transformer drops performance with the addition of solarization and Gaussian blur. Interestingly, the two arms converged with the use of Gaussian blur.}
\label{augmentaions}
\end{table*}

\subsection{Optimization}

In CASS we use Adam optimizer for both CNN and Transformer. This is a shift from the traditional use of SGD or stochastic gradient descent for CNNs. In this Table \ref{optimizer} we report the performance of CASS trained CNN and Transformer with the CNN using SGD and Adam optimizer. We observed that while the performance of CNN remained almost constant, the performance of Transformer dropped by almost 6\% with CNN using SGD.

\begin{table}[!htb]
\centering
\begin{tabular}{cll}
\hline
Optimiser for CNN                                   & CNN F1 Score    & Transformer F1 Score \\
\hline
Adam                                                & 0.8650±0.0001   & 0.8894±0.005         \\
SGD                                                 & 0.8648±0.0005   & 0.82355±0.0064       \\
\hline
\end{tabular}
\caption{We report the F1 metric of CASS trained with
a different set of optimizers for the CNN arm for 100 epochs. While there is no change in CNN's performance, the Transformer's performance drops around 6\% with SGD.}
\label{optimizer}
\end{table}

\subsection{Change in architecture}

\subsubsection{Changing Transformer and keeping the CNN same}

From Table \ref{differentTrasnformer} and \ref{differentTrasnformer_results}, we observed that CASS trained ViT Transformer with the same CNN consistently gained approximately 4.7\% over its
supervised counterpart. Furthermore, from Table \ref{differentTrasnformer_results} we observed that although ViT L/16 performs better than ViT B/16 on ImageNet ( \cite{rw2019timm}'s results), we observed that the trend is opposite on the autoimmune dataset. Hence, the supervised performance of architecture must be considered before pairing it with CASS.

\begin{table}[!htb]
\centering
\begin{tabular}{lll}
\hline
 Transformer            & CNN F1 Score  & Transformer F1 Score \\
\hline
 ViT Base/16   & 0.8650±0.001 & 0.8894± 0.005       \\
ViT Large/16  & 0.8481±0.001 & 0.853±0.004       \\
\hline
\end{tabular}
\caption{In this table we show performance of CASS for ViT large/16 with ResNet-50 and ViT base/16 with ResNet-50. We observed that CASS trained Transformers on average performed 4.7\% better than their supervised counterparts.}
\label{differentTrasnformer}
\end{table}

\begin{table}[!htb]
\centering
\begin{tabular}{ll}
\hline
\multicolumn{1}{c}{Architecture} & Testing F1 Score \\
\hline
ResnNet-50                       & 0.83895±0.007     \\
ViT Base/16                      & 0.8420±0.009     \\
ViT large/16                     & 0.80495±0.0077 \\
\hline
\end{tabular}
\caption{Supervised performance of ViT family on the autoimmune dataset. We observed that as opposed to ImageNet performance, ViT large/16 performs worse than ViT Base/16 on the autoimmune dataset.}
\label{differentTrasnformer_results}
\end{table}

\subsubsection{Changing CNN and keeping the Transformer same}

Table \ref{sameTransformer_atm} and \ref{cnn_atm} we observed that similar to changing Transformer while keeping CNN same, CASS trained CNNs gained an average of 3\% over their supervised counterparts. For ResNet-200 \cite{rw2019timm} doesn't have ImageNet initialization hence used random initialization. 

\begin{table*}[!htb]
\centering
\begin{tabular}{llll}
\hline
CNN                           & Transformer                  & \multicolumn{2}{l}{100\% Label Fraction} \\
                              &                              & CNN F1 score     & Transformer F1 score   \\
\hline
\multicolumn{1}{c}{ResNet-18 (\textbf{11.69M})} & \multirow{3}{*}{ViT Base/16 (\textbf{86.86M})} & 0.8674±4.8e-5     & 0.8773±5.29e-5           \\
ResNet-50 (\textbf{25.56M})                     &                              & 0.8680±0.001   & 0.8894± 0.0005         \\
ResNet-200 (\textbf{64.69M})                    &                              & 0.8517±0.0009     & 0.874±0.0006     \\
\hline

\end{tabular}
\caption{F1 metric comparison between the two arms of CASS trained over 100 epochs, following the protocols and procedure listed in Appendix E. The numbers in parentheses show the parameters learned by the network. We use \cite{rw2019timm} implementation of CNN and transformers, with ImageNet initialisation except for ResNet-200.}
\label{sameTransformer_atm}
\end{table*}

\begin{table}[!htb]
\centering
\begin{tabular}{ll}
\hline
\multicolumn{1}{c}{Architecture} & Testing F1 Score                                        \\
\hline
ResnNet-18                       & 0.8499±0.0004                                           \\
ResnNet-50                       & 0.83895±0.007                                            \\
ResnNet-200                      & 0.833±0.0005  \\
\hline
\end{tabular}
\caption{Supervised performance of the ResNet CNN family on the autoimmune dataset.}
\label{cnn_atm}
\end{table}

\section{Additional Results}

We conducted some further experimentation to check for collapse and ablation studies. The model did not report collapse in any case. The following results are calculated following the protocols in Appendix E on the brain MRI classification dataset.

\subsection{Changing Batch Size}

This section presents the results of varying the batch size in the brain MRI classification dataset. In the standard implementation of CASS, we used a batch size of 16; here, we showed results for batch sizes 8 and 32. The largest batch size we could run was 34 on a single GPU of 48 GB video memory. Hence 32 was the biggest batch size we showed in our results. We present these results in Table \ref{batchsize_bmri}. However, the performance of CNN remained nearly constant with a change of less than a percentage. The performance of the Transformer drops as we increase the batch size. Since CASS was developed with the requirements of running on small datasets, with an overall size smaller than the batch size of the current state of the art techniques, its peak performance for small batch size justifies its development. Furthermore, from Table \ref{batchsize_atm} and  Section 5.2, we observed that Transformer is the better performing architecture out of the two and that with batch size change, it was unaffected. Similarly, from Table \ref{batchsize_bmri} and Section 5.4, we observed that CNN is the better performing architecture, and with batch size change, its performance is unaffected. Hence, we conclude that batch size change does not affect the leading architecture on a given dataset. 

\begin{table}[!htb]
\centering
\begin{tabular}{lll}
\hline
Batch Size & CNN F1 Score                       & Transformer F1 Score \\
\hline
8          & 0.9895±0.0025                      & 0.93158±0.0109       \\
16         & \multicolumn{1}{c}{0.9909± 0.0032} & 0.9279± 0.0213       \\
32         & 0.9848±0.011                       & 0.9176±0.006        \\
\hline
\end{tabular}
\caption{This table represents the results for different batch sizes on the brain MRI classification dataset.  We maintain the downstream batch size constant for all the three self-supervised batch sizes, following the standard experimental setup as mentioned in Appendix E. These results are for 100\% label fraction.}
\label{batchsize_bmri}
\end{table}

\subsection{Effect of the Number of Training Epochs}

We saw that there was an incremental gain in performance as we increased the number of self-supervised training epochs. For the opposite scenario, there wasn't a steep drop in performance like the existing self-supervised techniques when we reduce the number of self-supervised epochs. Table \ref{epochs_bmri} displays results for this experimentation. 
\begin{table}[!htb]
\centering
\begin{tabular}{lll}
\hline
Epochs & CNN F1 Score                       & Transformer F1 Score \\
\hline
50                              & 0.9795±0.0109                      & 0.9262±0.0181       \\
100                             & \multicolumn{1}{c}{0.9909± 0.0032} & 0.9279± 0.0213       \\
200                             & 0.9864±0.008                       & 0.9476±0.0012        \\
300                             & 0.9920±0.001                       & 0.9484±0.017   \\
\hline
\end{tabular}
\caption{ Performance comparison over varied number of epochs, from 50 to 300 epochs, the downstream training procedure and the CNN-transformer combination is kept constant across all the four experiments, only the number of self-supervised epochs has been changed.}
\label{epochs_bmri}
\end{table}

\subsection{Effect of Changing the ViT/CNN branch}

\subsubsection{Changing CNN while keeping Transformer same}
For this experiment we use ResNet family of CNNs along with ViT base/16 as our Transformer. We use ImageNet initialization for ResNet 18 and 50, while random initialization for ResNet-200. We present these results in Table \ref{differentcnn_bmri}. We observed that increase in performance of ResNet correlates to increase in performance of Transformer, hence implying that there is information transfer between the two.

\begin{table*}[!htb]
\centering
\begin{tabular}{llll}
\hline
CNN                           & Transformer                  & \multicolumn{2}{l}{100\% Label Fraction} \\
                              &                              & CNN F1 score     & Transformer F1 score   \\
\hline
\multicolumn{1}{c}{ResNet-18 (\textbf{11.69M})} & \multirow{3}{*}{ViT Base/16 (\textbf{86.86M})} & 0.9913±0.002     & 0.9801±0.007           \\
ResNet-50 (\textbf{25.56M})                     &                              & 0.9909±0.0032    & 0.9279± 0.0213         \\
ResNet-200 (\textbf{64.69M})                    &                              & 0.9898±0.005     & 0.9276±0.017     \\
\hline

\end{tabular}
\caption{F1 metric comparison between the two arms of CASS trained over 100 epochs, following the protocols and procedure listed in Appendix E. The numbers in parentheses show the parameters learned by the network. We use \cite{rw2019timm} implementation of CNN and transformers, with ImageNet initialisation except for ResNet-200.}
\label{differentcnn_bmri}
\end{table*}

\subsubsection{Changing Transformer while keeping CNN same}
For this experiment we keep the CNN as constant and study the effect of changing the Transformer. For this experiment we use ResNet as our choice of CNN and ViT base and large Transformers with 16 patches. Additionally we also report performance for DeiT-B with ResNet-50. We report these results in Table \ref{transformer_bmri}. Similar to Table \ref{differentTrasnformer} we observe that changing Transformer from ViT Base to Large while keeping the number of tokens same at 16, performance drops. Additionally, for approximately the same size, out of DEiT base and ViT base Trasnformers, DEiT performs much better than ViT base. 

\begin{table*}[t]
\centering
\begin{tabular}{clll}
\hline
CNN                                                                           & Transformer            & CNN F1 Score  & Transformer F1 Score \\
\hline
\multirow{3}{*}{\begin{tabular}[c]{@{}c@{}}ResNet-50\\ (25.56M)\end{tabular}} & DEiT Base/16 (86.86M)     & 0.9902±0.0025   & 0.9844±0.0048     \\
& ViT Base/16 (86.86M)   & 0.9909±0.0032 & 0.9279± 0.0213       \\
                                                                              & ViT Large/16 (304.72M) & 0.98945±2.45e-5 & 0.8896±0.0009       \\
                                                                              
\hline
\end{tabular}
\caption{For the same number of Transformer parameters, DEiT-base with ResNet-50 performed much better than ResNet-50 with ViT-base. The difference in their CNN arm is ~0.10\%. On ImageNet DEiT-base has a top1\% accuracy of 83.106 while ViT-base has an accuracy of 86.006. We use both the Transformers with 16 patches. [ResNet-50 has an accuracy of 80.374] }
\label{transformer_bmri}
\end{table*}

\subsubsection{Using CNN in both arms}

Until now we have experimented by using a CNN and a Transformer in CASS. In this section we present results for using two CNNs in CASS. We pair ResNet-50 with DenseNet-161. We observe that both the CNNs fail to reach the benchmark set by ResNet-50 and ViT-B/16 combination. Although training the ResNet-50-DenseNet-161 pair takes 5 hour 24 minutes which is less than the 7 hours 11 minutes taken by the ResNet-50-ViT-B/16 combination to be trained with CASS. We compare these results in Table \ref{bothcnn}.

\begin{table*}[!htb]
\centering
\begin{tabular}{clll}
\hline
CNN                        & \begin{tabular}[c]{@{}l@{}}Architecture in\\  arm 2\end{tabular} & F1 Score of ResNet-50 arm  & F1 Score of arm 2 \\
\hline
\multirow{2}{*}{ResNet-50} & ViT Base/16                                                              & 0.9909±0.0032 & 0.9279± 0.0213            \\
                           & DenseNet-161                                                             & 0.9743±8.8e-5 & 0.98365±9.63e-5    \\
\hline
\end{tabular}
\caption{We observed that for the ResNet-50-DenseNet-161 pair, we train 2 CNNs instead of 1 in our standard setup of CASS. Furthermore, none of these CNNs could match the performance of ResNet-50 trained with the ResNet-50-ViT base/16 combination. Hence, by adding a Transformer-CNN combination, we transfer information between the two architectures that would have been missed otherwise.}
\label{bothcnn}
\end{table*}

\subsubsection{Using Transformer in both arms}

Similar, to the above section, for this section we use a Transformer-Transformer combination instead of a CNN-Transformer combination. For this we use Swin-Transformer patch-4/window-12 \cite{Liu_2021_ICCV} alongside ViT-B/16 Transformer. We observe that the performance for ViT/B-16 improves by around ~1.3\% when we use Swin Transformer. Although this comes at computational cost. Swin-ViT combination took 10 hours to train as opposed to 7 hours 11 minutes taken by the ResNet-50-ViT-B/16 combination to be trained with CASS. Even with the increase in time taken to train the Swin-ViT combination, it is still almost 50\% less than DINO.
We present these results in Table \ref{bothT}.

\begin{table*}[!htb]
\centering
\begin{tabular}{clll}
\hline
\begin{tabular}[c]{@{}c@{}}Architecture in\\  arm 1\end{tabular} & Transfomer                   & F1 Score of arm 1  & F1 Score of ViT-B/16 arm \\
\hline
\multicolumn{1}{l}{ResNet-50}                                            & \multirow{2}{*}{ViT Base/16} & 0.9909±0.0032  & 0.9279± 0.0213            \\
Swin Tranformer                                                          &                              & 0.9883±1.26e-5 & 0.94±8.12e-5     \\
\hline
\end{tabular}
\caption{We present the results for using Transformers in both the arms and compare the results with CNN-Transformer combination.}
\label{bothT}
\end{table*}

\subsection{Effect of Initialization}

We use ImageNet initialized CNN and Transformers for CASS, DINO and supervised training. We use Timm's library for these initialization \cite{rw2019timm}. ImageNet initialization is preferred for transfer learning in medical image analysis not because of feature reuse but becuase ImageNet weights allow for faster convergence through better weight scaling \cite{raghu2019transfusion}. But sometimes pretrained weights might be hard to find, so we study CASS' performance with random and ImageNet initialization in this section.
We observed that performance almost remained the same with minor gains when the initialization was altered for the two networks. Table \ref{init} presents the results for this experimentation.

\begin{table}[!htb]
\centering
\begin{tabular}{lll}
\hline
Initialisation & CNN F1 Score  & Transformer F1 Score \\
\hline
Random        & 0.9907±0.009  & 0.9316±0.027         \\
Imagenet      & 0.9909±0.0032 & 0.9279± 0.0213      \\
\hline
\end{tabular}
\caption{We observe that the Transformer gains some performance with random initialization, although performance has more variance when used with random initialization.}
\label{init}
\end{table}

\subsection{Using softmax and sigmoid layer in CASS}

As noted in Fig 1 and Section 3.1, CASS doesn’t use a softmax layer like DINO (\cite{caron2021emerging}) before computing loss. The output logits of the two networks have been used to combine the two architectures in a response-based knowledge distillation (\cite{gou2021knowledge}) manner instead of using soft labels from the softmax layer. In this section, we study the effect of using an additional softmax layer on CASS. Furthermore, we also study the effect of adding a sigmoid layer instead of a softmax layer and compare it with a CASS model that doesn’t use the sigmoid or the softmax layer. We present these results in Table \ref{softmax}. We observed that not using sigmoid and softmax layers in CASS yield the best result for both CNN and Transformers.

\begin{table}[!htb]
\centering
\begin{tabular}{lll}
\hline
Techniques & CNN F1 Score  & Transformer F1 Score \\
\hline
Without Sigmoid or Softmax       & 0.8650±0.0001  & 	0.8894±0.005         \\
With Sigmoid Layer      & 0.8296±0.00024 & 0.8322±0.004      \\
With Softmax Layer      & 0.8188±0.0001 & 0.8093±0.00011      \\
\hline
\end{tabular}
\caption{We observe that performance reduces when we introduce sigmoid or softmax layer.}
\label{softmax}
\end{table}
\section{Result Analysis}

\subsection{Time complexity analysis}
In section 5.1, we observed that CASS takes 69\% less time as compared to DINO. This reduction in time could be attributed to the follwoing reasons:
\begin{enumerate}
    \item In DINO, augmentations are applied twice as opposed to just once in CASS. Furthermore, we per application CASS uses less number of augmentations as compared to DINO.
    \item Since the architectures, used are different, there is no scope of parameter sharing between them. A major chunk of time is saved in just updating the two architectures, after each epoch instead of re-initializing architectures with lagging parameters.  
\end{enumerate}

To qualitatively understand the effect of training CNN with 
Transformer, we study the feature maps of CNN and attention maps of Transformers trained using CASS and supervised techniques. To reinstate, based on the study by \cite{Raghu2021DoVT} since CNN and Transformer extract different kinds of features from the same input, combing the two of them would help us create positive pairs for self-supervised learning. In doing so, we would transfer between the two architectures, which is not innate. We have already seen that this yield better performance in most cases over four different datasets and with three different label fractions. In this section, we study this gain qualitatively with the help of feature maps and class attention maps. 

\subsection{Feature maps}
In this section, we study the feature maps from the first five layers of the ResNet-50 model trained with CASS and supervision. We extracted feature maps after the Conv2d layer of ResNet-50. We present the extracted features in Figure \ref{fig:fmaps}. We observed that CASS-trained CNN could retain much more detail about the input image than supervised CNN.

\begin{figure}
    \centering
    \includegraphics[width=0.5\linewidth]{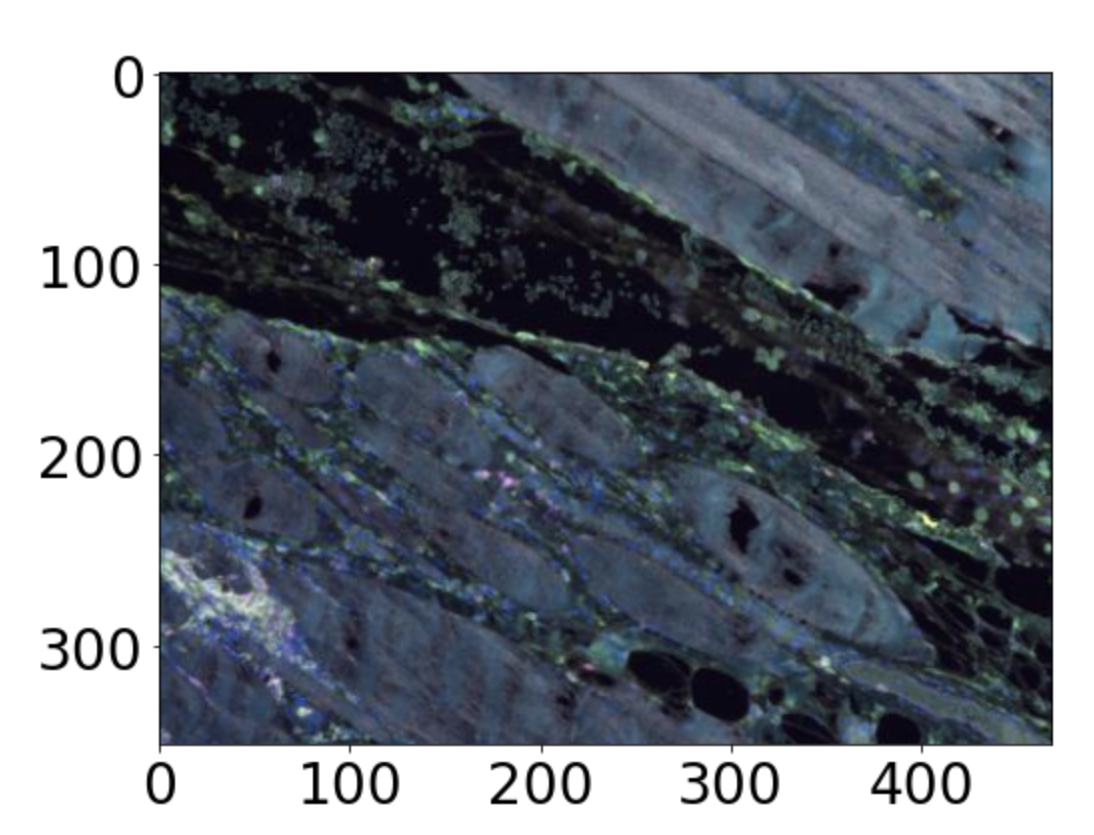}
    \caption{Sample image used from the test set of the autoimmune dataset.}
    \label{fig:sample_image}
\end{figure}

\begin{figure*}[t]
    \includegraphics[width=0.9\linewidth]{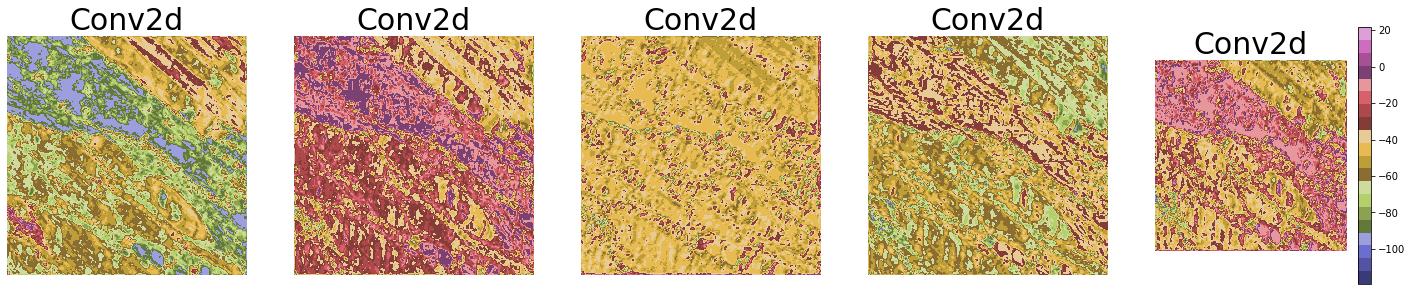}
    \includegraphics[width=0.9\linewidth]{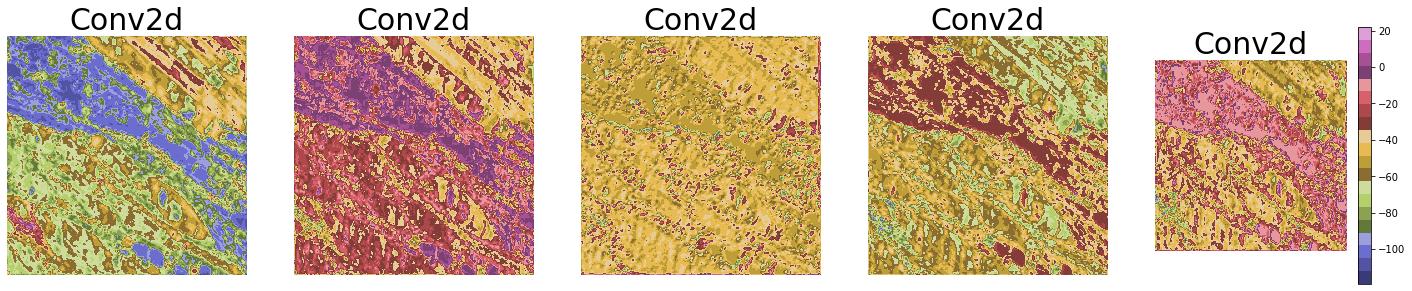}
    \caption{ At the top we have features extracted from top 5 layers of supervised ResNet-50 while at the bottom we have features extracted from top 5 layers of CASS trained ResNet-50. We supplied both the networks with the same input ( shown in Figure~\ref{fig:sample_image}). }
    \label{fig:fmaps}
\end{figure*}

\subsection{Class attention maps}

We have already studied the class attention maps over a single image in Section 5.6. This section will study the average class attention maps for all four datasets. We studied the attention maps averaged over 30 random samples for autoimmune, dermofit, and brain MRI datasets. Since the ISIC 2019 dataset is highly unbalanced, we averaged the attention maps over 100 samples so that each class may have an example in our sample. We maintained the same distribution as the test set, which has the same class distribution as the overall training set. We observed that CASS-trained Trasformers were better able to map global and local attention maps due to Transformers ability to map global dependencies and by learning features sensitive to translation equivariance and locality from CNN.

\subsubsection{Autoimmune dataset}
\begin{figure}[!ht]
    \centering
    \includegraphics[width=0.6\linewidth]{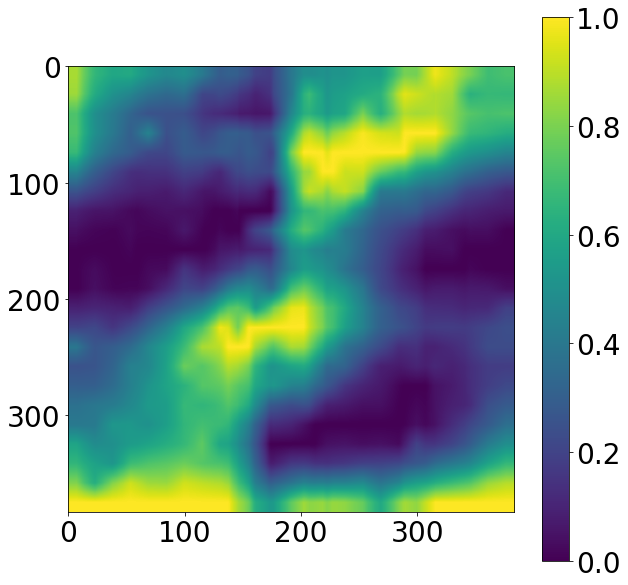}
    \includegraphics[width=0.6\linewidth]{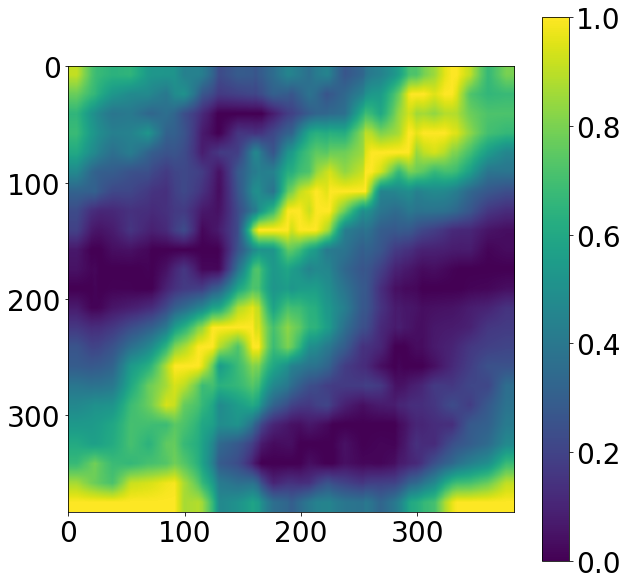}
   
    \caption{To ensure the consistency of our study, we studied average attention maps over 30 sample images from the autoimmune dataset. The top image is the overall attention map averaged over 30 samples for supervised Transformer, while the one at the bottom is for CASS trained Transformer.}
    \label{fig:colton_avg}
\end{figure}

We study the class attention maps averaged over 30 test samples for the autoiimune dataset in Figure \ref{fig:colton_avg}. We observed that the CASS-trained Transformer has much more attention in the center as compared to the supervised Transformer. This extra attention could be attributed that a Transformer on its own couldn't map out is due to the information transfer from CNN. Another, feature to observe is that the attention map of CASS-trained Transformer is much more connected than that of supervised Transformer.

\subsubsection{Dermofit dataset} We present the average attention maps for the dermofit dataset in Figure \ref{fig:dermofit_attn}. We observed that the CASS-trained Transformer is able to pay a lot more attention to the center part of the image. Furthermore, the attention map of CASS-trained Transformer is much more connected as compared to the supervised Transformer. So, overall with CASS, the Transformer is not only able to map long-range dependencies which are innate to Transformers but is also able to make more local connections with the help of features sensitive to translation equivariance and locality from CNN.

\subsubsection{Brain tumor MRI classification dataset} We present the results for the average class attention maps in Figure \ref{fig:bmri_attn}. We observed that a CASS-trained Transformer could better capture long and short-range dependencies than a supervised Transformer. Furthermore, we observed that a CASS-trained Transformer's attention map is much more centered than a supervised Transformer's. From Figure \ref{fig:bmri} we can observe that most MRI images are center localized, so having a more centered attention map is advantageous in this case.   

\subsubsection{ISIC 2019 dataset} The ISIC-2019 dataset is one of the most challenging datasets out of the four datasets. ISIC 2019 consists of images from the HAM10000 and BCN\_20000 datasets \cite{cassidy2022analysis,Gessert2020SkinLC}. For the HAM1000 dataset, it is difficult to classify between 4 classes (melanoma and melanocytic nevus), (actinic keratosis and benign keratosis). HAM10000 dataset contains images of size 600×450, centered and cropped around the lesion. Histogram corrections have been applied to only a few images. The BCN\_20000 dataset contains images of size 1024×1024. This dataset is particularly challenging as many images are uncropped, and lesions are in difficult and uncommon locations. Hence, in this case, having more spread-out attention maps would be advantageous instead of a more centered one. From Figure \ref{fig:isic_attn}, we observed that CASS-trained Transformer has a lot more spread attention map than a supervised Transformer. Furthermore, CASS-trained Transformer is also able to attend the corners far better than supervised Transformer.

\section{Expansion on experimentation details}

\subsection{Datasets}

\subsubsection{Choice of Datasets} We chose four medical imaging datasets with diverse sample sizes ranging from 198 to 25,336 and diverse modalities to study the performance of existing self-supervised techniques and CASS.
Most of the existing self-supervised techniques have been studied on million image datasets, but medical imaging datasets, on average, are much smaller than a million images. We expand this to include datasets of emerging and underrepresented diseases with only a few hundred samples, like the autoimmune dataset in our case (198 samples). To the best of our knowledge, no existing literature studies the effect of self-supervised learning on such a small dataset. Furthermore, we chose the dermofit dataset because all the images are taken using an SLR camera, and no two images are the same size. Image size in dermofit varies from 205×205 to 1020×1020. MRI images constitute a large part of medical imaging; hence we included this dataset in our study. So, to incorporate them in our study, we included the Brain tumor MRI classification dataset. Furthermore, it is our study's only black and white dataset; the other three datasets are RGB.
The ISIC 2019 is a unique dataset as it contains multiple pairs of hard-to-classify classes (Melanoma - melanocytic nevus and actinic keratosis - benign keratosis) and different image sizes - out of which only a few have been prepossessed. It is a highly imbalanced dataset containing samples with lesions in difficult and uncommon locations.



    \begin{figure}[!h]
    \centering
    \includegraphics[width=0.4\linewidth]{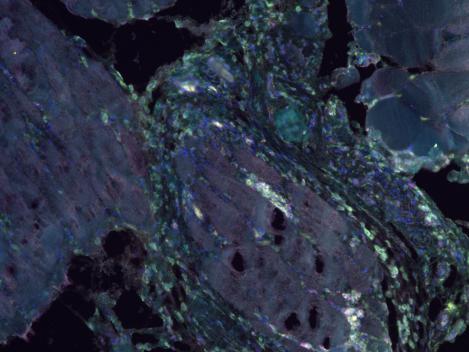}
    \includegraphics[width=0.4\linewidth]{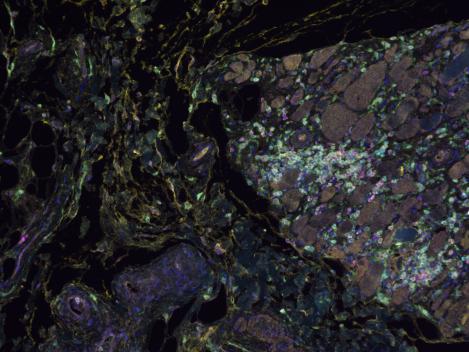}
    \caption{Sample of autofluorescence slide images from the muscle biopsy of patients with dermatomyositis - a type of autoimmune disease.}
    \label{fig:autoimmune}
\end{figure}

    
    \begin{figure}[!ht]
    \centering
    \includegraphics[width=0.45\linewidth]{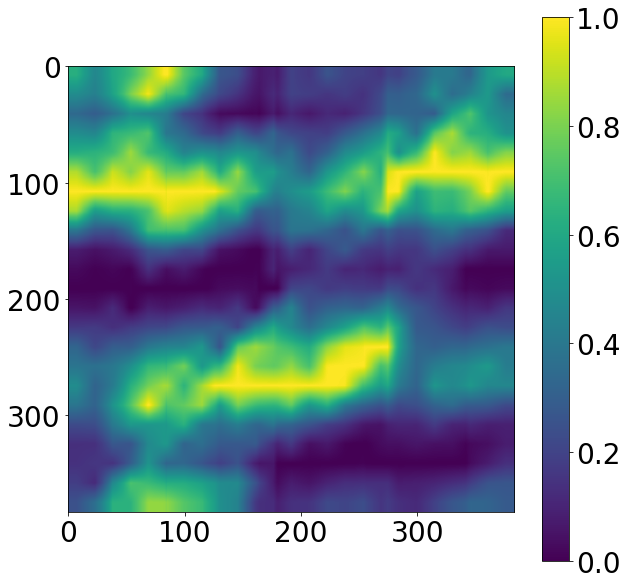}
    \includegraphics[width=0.45\linewidth]{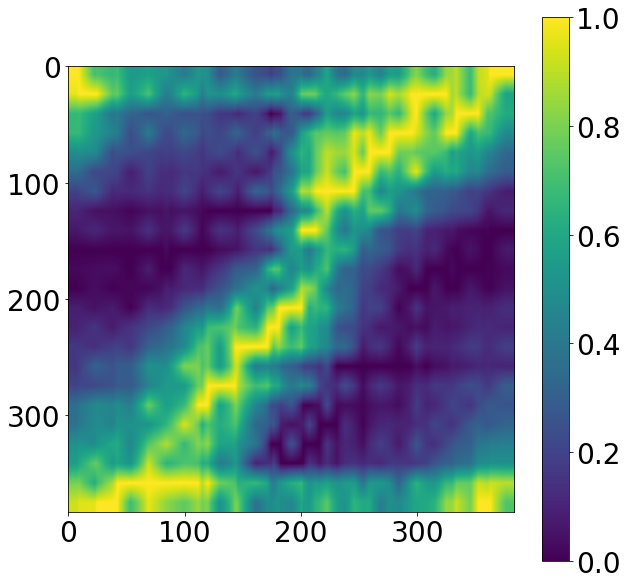}
    
    \caption{Class attention maps averaged over 30 samples of the dermofit dataset for supervised Transformer (on the left), and CASS trained Transformer (on the right). }
    
    \label{fig:dermofit_attn}
\end{figure}
    
    \begin{figure}[!htb]
    \centering  
    \includegraphics[width=0.3\linewidth]{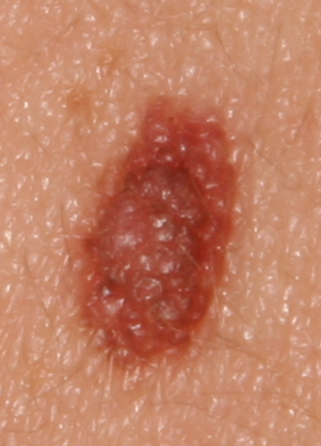}
    \includegraphics[width=0.3\linewidth]{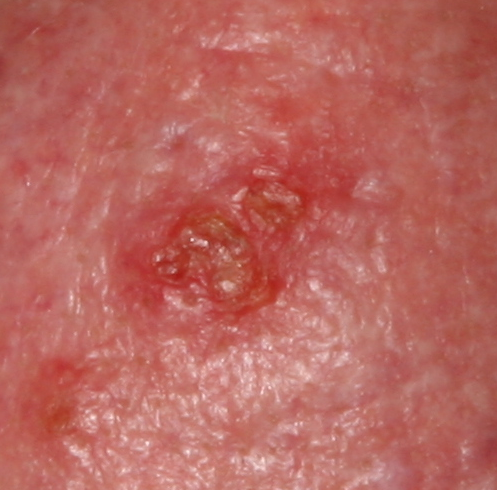}
    \includegraphics[width=0.3\linewidth]{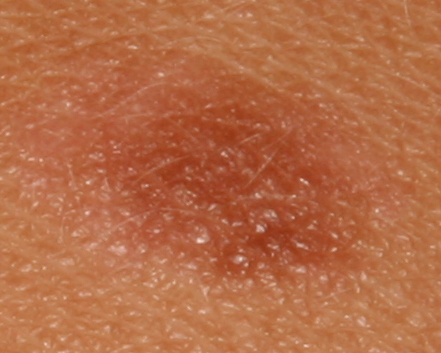}
     \hspace{1cm}
    \caption{Sample images from the Dermofit dataset.}
    \label{fig:dermofit}
    \end{figure}

    
     \begin{figure}[!htb]
     \centering
    \includegraphics[width=0.25\linewidth]{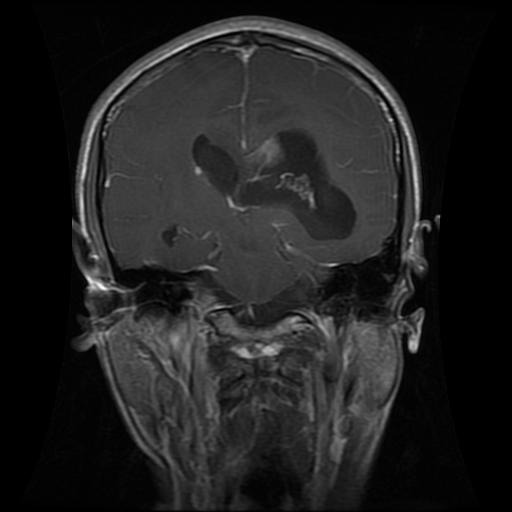}
    \includegraphics[width=0.25\linewidth]{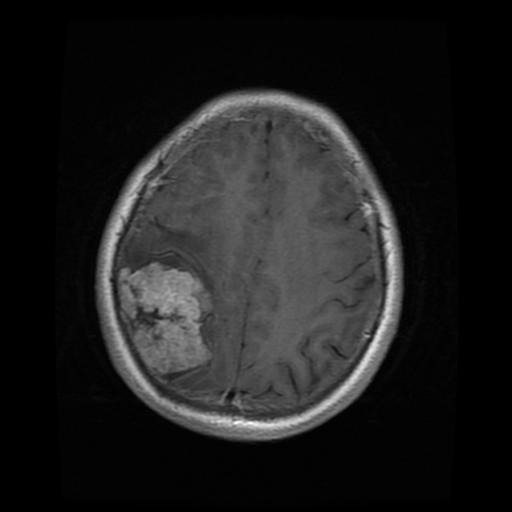}
    \includegraphics[width=0.25\linewidth]{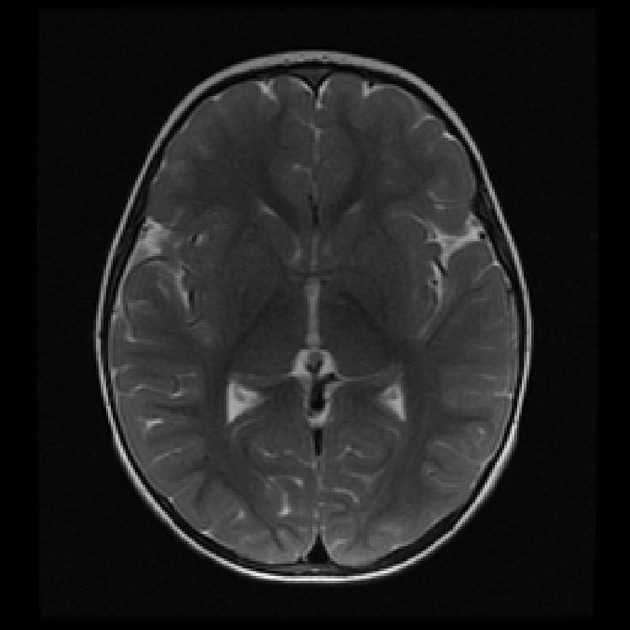}
     \hspace{1cm}
    \caption{Sample images of brain tumor MRI dataset, Each image corresponds to a prediction class in the data set glioma (Left), meningioma (Center) and No tumor (Right)  }
    \label{fig:bmri}
    \end{figure}
    
    \begin{figure}[!ht]
    \centering
    \includegraphics[width=0.45\linewidth]{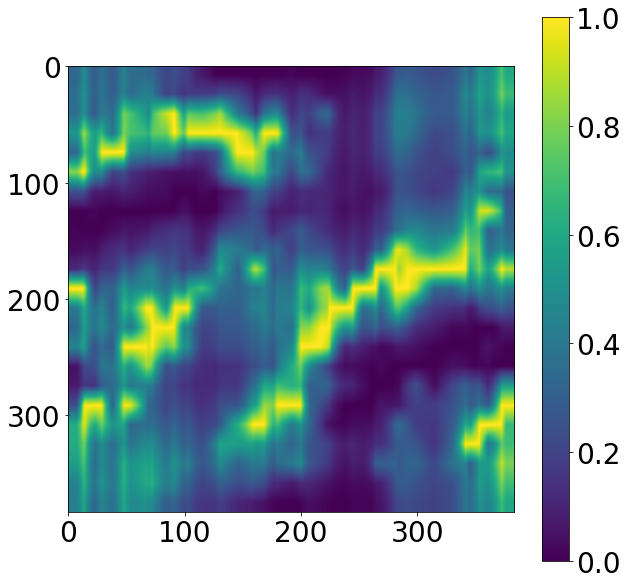}
    \includegraphics[width=0.45\linewidth]{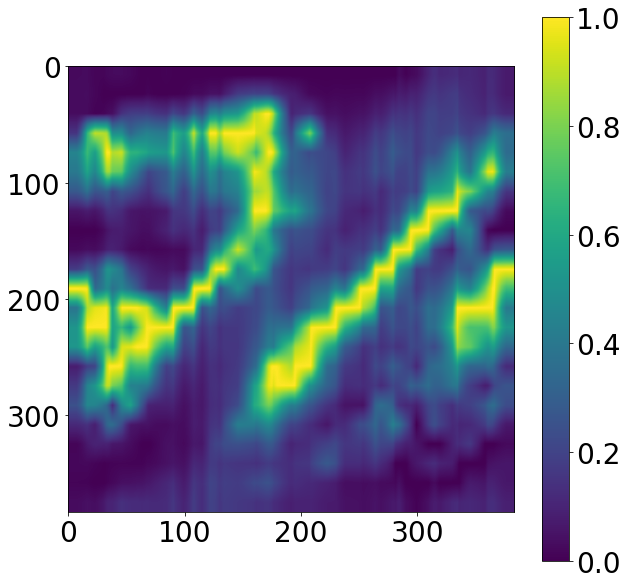}
    \caption{Class attention maps averaged over 30 samples of the brain tumor MRI classification dataset for supervised Transformer (on the left) and CASS trained Transformer (on the right).}
    
    \label{fig:bmri_attn}
\end{figure}
    
    
    \begin{figure}[!ht]
    \centering
    \includegraphics[width=0.45\linewidth]{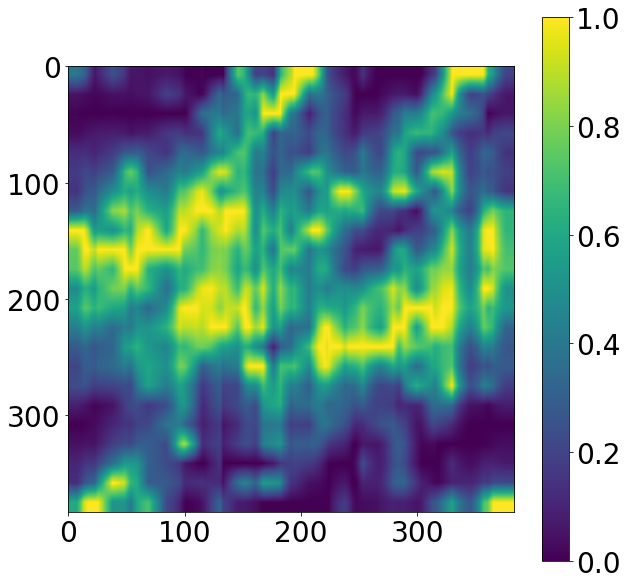}
    \includegraphics[width=0.45\linewidth]{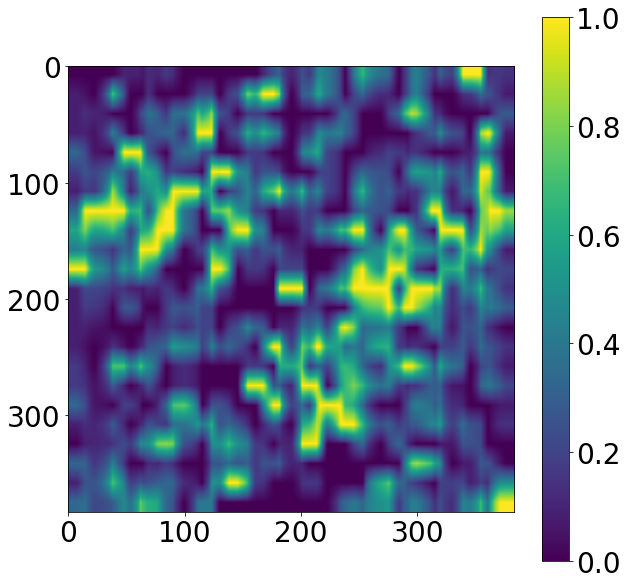}
    \caption{Class attention maps avergaed over 100 samples form the ISIC-2019 dataset for supervised Transformer (on the left) and CASS trained Transformer (on the right).}
    
    \label{fig:isic_attn}
\end{figure}
    
    \begin{figure}[!htb]
    \centering  
    \includegraphics[width=0.25\linewidth]{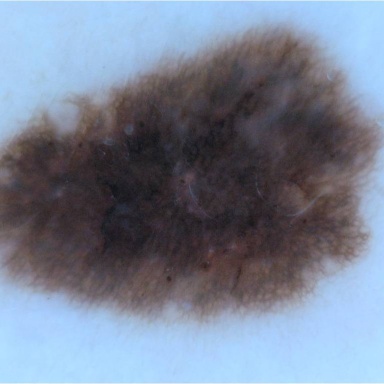}
    \includegraphics[width=0.25\linewidth]{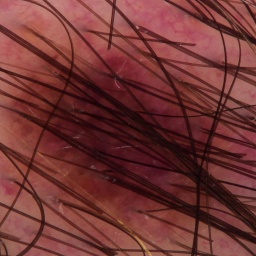}
    \includegraphics[width=0.25\linewidth]{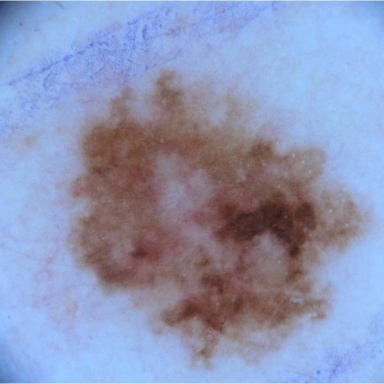}
     \hspace{1cm}
    \caption{Sample images from the ISIC-2019 challenge dataset.}
    \label{fig:isic}
    \end{figure}

\subsection{Self-supervised training}

\subsubsection{Protocols}
\begin{itemize}
    \item Self-supervised learning was only done on the training data and not on the validation data. We used \url{https://github.com/PyTorchLightning/pytorch-lightning} to set the pseudo-random number generators in PyTorch, NumPy and (python.random).

   \item We run training over five different seed values, and report mean results with variance in each table. We don't perform a seed value sweep to extract anymore performance \cite{picard2021torch}.
   
   \item For DINO implementation we use Phil Wang's implementation: \url{https://github.com/lucidrains/vit-pytorch}.
   
   \item For implementation of CNNs and Transformers we use timm's library \cite{rw2019timm}.
   
   \item For all experiments, ImageNet~\cite{deng2009imagenet} initialised CNN and Transformers were used.
\end{itemize}

\subsubsection{Augmentations}

\begin{itemize}

\item Resizing: Resize input images to 384×384 with bilinear interpolation.
\item Color jittering: change the brightness, contrast, saturation and hue of an image or apply random perspective with a given probability. We set the degree of distortion to 0.2 (between 0 and 1) and use bilinear interpolation, with an application probability of 0.3.
\item Color jittering or apply random affine transformation of the image keeping center invariant with degree 10, with an application probability of 0.3.
\item Horizontal and Vertical flip. Each with an application probability of 0.3.
\item Channel normalisation with mean (0.485, 0.456, 0.406) and standard deviation (0.229, 0.224, 0.225).

\end{itemize}

\subsubsection{Hyper-parameters}

\begin{itemize}
    \item Optimization: We use stochastic weighted averaging over Adam optimiser with learning rate (LR) set to 1e-3 for both CNN and vision transformer (ViT). This is a shift from SGD which is usally used for CNNs.
    
    \item Learning Rate: Cosine annealing learning rate is used with 16 iterations and a minimum learning rate of 1e-6. Unless mentioned otherwise, this setup was trained over 100 epochs. These were then used as initialisation for the downstream supervised learning. The standard batch size is 16.
    
\end{itemize}

\subsection{Supervised training}

\subsubsection{Augmentations}
We use the same set of augmentations used in self-supervised training.
\subsubsection{Hyper-parameters}
\begin{itemize}
    \item We use Adam optimiser with lr set to 3e-4 and a cosine annealing learning schedule.
    \item Since, all medical datasets have class imbalance we address it by using focal loss \cite{Lin2017FocalLF} as our choice of loss function with the alpha value set to 1 and the gamma value to 2. In our case it uses minimum-maximum normalised class distribution as class weights for focal loss.
    \item We train for 50 epochs. We also use a five epoch patience on validation loss to check for early stopping. This downstream supervised learning setup is kept the same for CNN and Transformers.
\end{itemize}

We repeat all the experiments five time with different seed values and then present the average results in all the tables.

\section{Miscellaneous}

\subsubsection{Choice of Self-supervised technique for benchmarking}
Self-supervised learning has seen tremendous growth in recent years, not only in computer vision but also with other applications in deep learning. Within computer vision, leveraging unlabeled data has removed a significant bottleneck. We have already touched upon these points in Section 2.2. This section aims to explain the reasoning the thought process of choosing the benchmarking self-supervised technique in our context. Since with CASS, we wanted to propose an alternate way of making positive pairs with architectural differences instead of differences in augmentation. For a fair comparison, we wanted to compare the performance of both CNN as well as Transformers and identify and evaluate performance differences. Most of the existing methods that researchers implement only address CNNs (\cite{Caron2020UnsupervisedLO,caron2021emerging}; \cite{Grill2020BootstrapYO} ; \cite{Chen2020ASF,zbontar2021barlow}), while other like MAE \cite{he2021masked} only focus on Transformers. Furthermore, we discarded the contrastive techniques using negative pairs as they are computationally intensive to run and do not fit in with our goal of creating an efficient self-supervised technique. Hence, the only self-supervised technique without high memory dependency and addressed for CNN as well as Transformers came out was DINO \cite{caron2021emerging}.

\subsubsection{Description of Metrics}
After performing downstream fine-tuning on the four datasets under consideration, we analyze the CASS, DINO, and Supervised approaches on specific metrics for each dataset. The choice of this metric is either from previous work or as defined by the dataset provider. For the Autoimmune dataset, Dermofit, and Brain MRI classification datasets based on previous works, we use the F1 score as our metric for comparing performance, which is defined as $F1 = \frac{2*Precision*Recall}{Precision+Recall} = \frac{2*TP}{2*TP+FP+FN}$

For the ISIC-2019 dataset, as mentioned by the competition organizers, we used the recall score as our comparison metric, which is defined as: $Recall = \frac{TP}{TP+FN}$

For the above two equations, TP: True Positive, TN: True Negative, FP: False Positive, and FN: False Negative.

\subsubsection{Calculating the gain in performance} 
To calculate the performance gain per label fraction, we sum the improvement over both CNN and Transformers, followed by dividing that by the number of the participating datasets. For example, for the 1\% label fraction gain, we divided by 2, as we only calculated results for 1\% label fraction using the 1) brain tumor MRI, and 2) ISIC-2019 dataset.


\end{document}